\documentclass{IEEEtran}
\usepackage{cite}
\usepackage{amsmath,amssymb,amsfonts}
\usepackage{algorithmic}
\usepackage{graphicx}
\usepackage{caption}
\usepackage{subfig}
\usepackage{textcomp}
\def\BibTeX{{\rm B\kern-.05em{\sc i\kern-.025em b}\kern-.08em
    T\kern-.1667em\lower.7ex\hbox{E}\kern-.125emX}}
\begin{document}
\title{Segmenting Ships in Satellite Imagery With Squeeze and Excitation U-Net}
\author{Venkatesh~R and Anand~Mehta
\thanks{Venkatesh R is with the Computer Science Department, SRM Institute of Science and Technology, Chennai, India
(e-mail: rvenkatesh\_mp@srmuniv.edu.in).}
\thanks{Anand Mehta is with 
the Computer Science Department, SRM Institute of Science and Technology, Chennai, India
(e-mail: anand.me@ktr.srmuniv.ac.in ).}
}

\maketitle

\begin{abstract}
The ship-detection task in satellite imagery presents significant obstacles to even the most state of the art segmentation models due to lack of labelled dataset or approaches which are not able to generalize to unseen images. The most common methods for semantic segmentation involve complex two-stage networks or networks which make use of a multi-scale scene parsing module. In this paper, we propose a modified version of the popular U-Net architecure called Squeeze and Excitation U-Net and train it with a loss that helps in directly optimizing the intersection over union (IoU) score. Our method gives comparable performance to other methods while having the additional benefit of being computationally efficient.
\end{abstract}

\begin{IEEEkeywords}
Segmentation, Geospatial, Computer Vision, Deep Learning, Satellite Imagery
\end{IEEEkeywords}

\section{Introduction}
\label{sec:introduction}
Automatic detection of objects of interests is a task that has been a challenge in the computer vision community for decades. A lot of work has been done over the past 10 years \cite{b2} to automatically extract objects from satellite imagery but with limited operational results. Often these solutions work with a limited dataset but are not able to generalize on unseen data. With the advent of deep learning algorithms which make use of convolutional neural networks (CNN's), there has been a lot of advancement in this field often producing state of the art results. 

Shipping traffic has grown rapidly over the last couple of decades. To handle illegal shipping and infractions at seas, maritime bodies usually monitor shipping traffic manually by going through each image. However, this work is a time-consuming task and requires qualified people. Automating this task is of significant importance to many. The advances in computer vision along with the availability of high-resolution data at a higher frequency will lead to the automation of these tasks.

\section{Related Work}
Deep learning has attracted a lot of attention, especially when applied to computer vision related tasks. Since AlexNet \cite{b4} was used to win the ImageNet challenge, CNN's have been applied to a variety of tasks. Fully-Convolutional Network \cite{b5} (FCN) was one of the first architectures for segmentation based on CNN's. Since then a number of architectures have evolved based on a similar structure. U-Net \cite{b17} which is a FCN having an encoder-decoder architecture along with skip-connections which was designed to perform in the absence of a large amount of data. In this paper, we make use of a modified version of U-Net by re-calibrating the learned feature maps using squeeze and excitation module \cite{b6}.

\begin{figure}%
    \centering
    \subfloat{{\includegraphics[width=2.7cm]{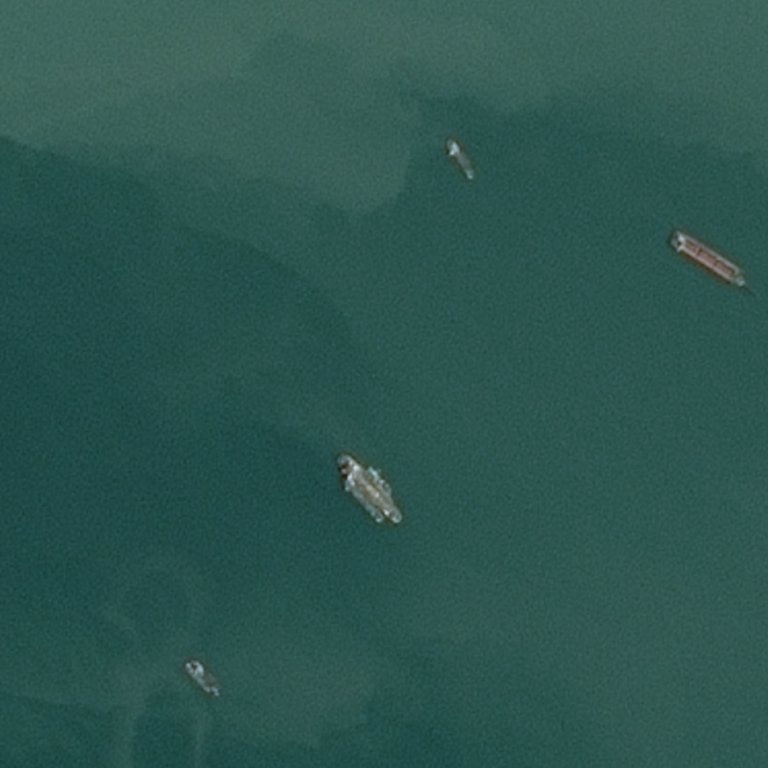} }}%
    \subfloat{{\includegraphics[width=2.7cm]{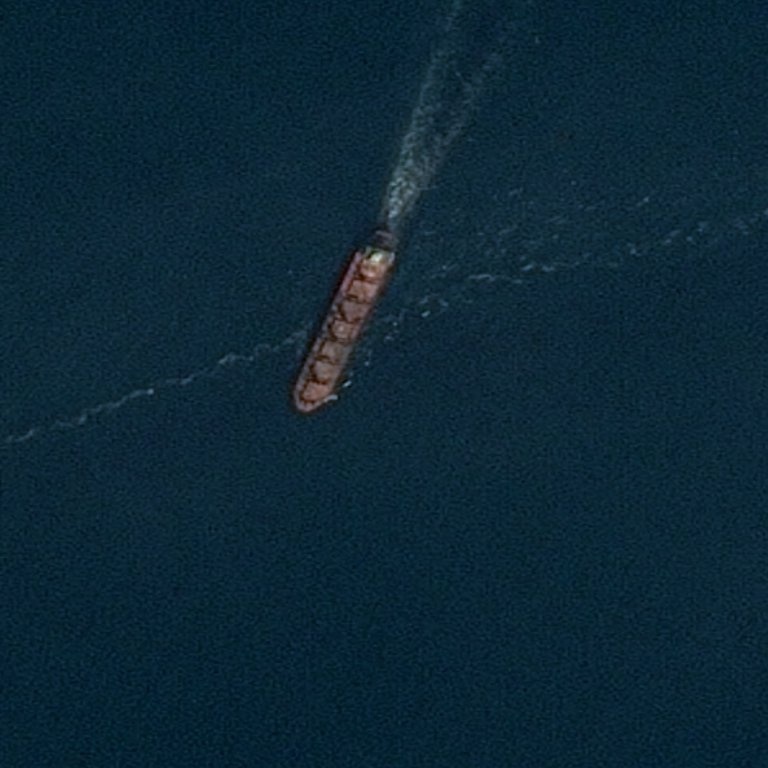} }}%
    \subfloat{{\includegraphics[width=2.7cm]{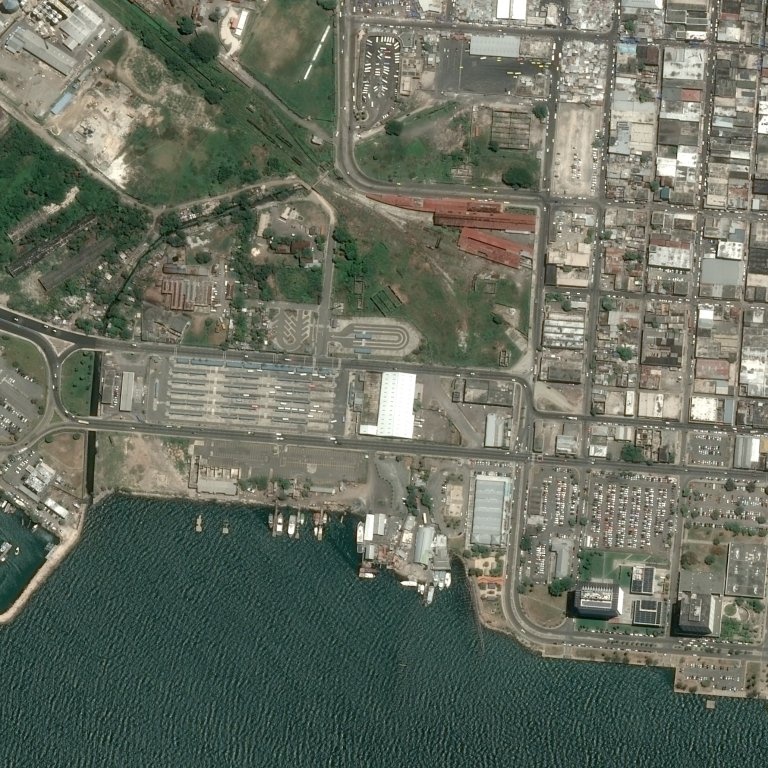} }}%
    \caption{Example images from the dataset}%
    \label{fig:examples}%
\end{figure}

When it comes to satellite images, Iglovikov et. al. \cite{b3} made use of a U-Net with a pretrained encoder (WideResnet-38) and recent improvements like activated batch normalization \cite{b15} which allows for memory savings and exponential linear unit \cite{b16} (ELU). 

Rakhlin et. al. \cite{b12} made use of U-net with an m46 encoder which they designed for saving memory. They trained their network using stochastic weight averaging (SWA) \cite{b13} which helps in finding much broader optima than gradient descent.

In our work we propose to use U-Net architecture with Resnet-34 \cite{b9} as the backbone and add spatial and channel squeeze and excitation (SE) blocks to our network for segmenting ships in satellite images. For optimizing our model we compare the performance of our model by using various loss functions. 

\section{Dataset and Evaluation Metric}
\begin{figure}
    \centering
    \includegraphics[width=7cm]{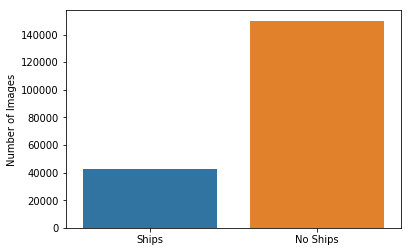}
    \caption{Images in each class (ships and no ships) vs. number of images}
    \label{fig:plot}
\end{figure}

The dataset for ship detection includes 3 channel RGB images.  Fig. \ref{fig:examples} shows examples of images from the dataset. The training set consists of 192,556 images and the validation set consists of 15,606 images. The size of images is 768x768. The dataset is highly imbalanced with close to 60\% of the images being empty with no masks. Fig. \ref{fig:plot} imbalance in the training set. The rest of the images contained masks of up to 15 ships in an image which includes ships of varying sizes.

The evaluation metric used for this task is $F\beta$-score at different IoU thresholds. The IoU score between a proposed set of objects and true objects is calculated as follows:
\begin{equation}
    IoU(A,B) = \frac{A \cap B}{A \cup B}
\end{equation}

The threshold for IoU ranges from 0.5 to 0.95 with a step size of 0.05. At each threshold value, the $F\beta$ score is calculated as follows:
\begin{equation}
    F_\beta(t) = \frac{(1 + \beta^2) . TP(t)}{(1 + \beta^2).TP(t) + \beta^2.FN(t) + FP(t)}
\end{equation}
Where TP, FP and FN denote true positives, false positives and false negatives and $\beta$ is set to 2. So it is equivalent to F2-score.
The final score is given by the mean of the F2-score at different IoU thresholds. 
\begin{equation}
    \frac{1}{|thresholds|}\sum_t F_2(t)
\end{equation}
The mean of F2 score is calculated as shown in (3).
\begin{figure*}
\centering
\includegraphics[width=\textwidth]{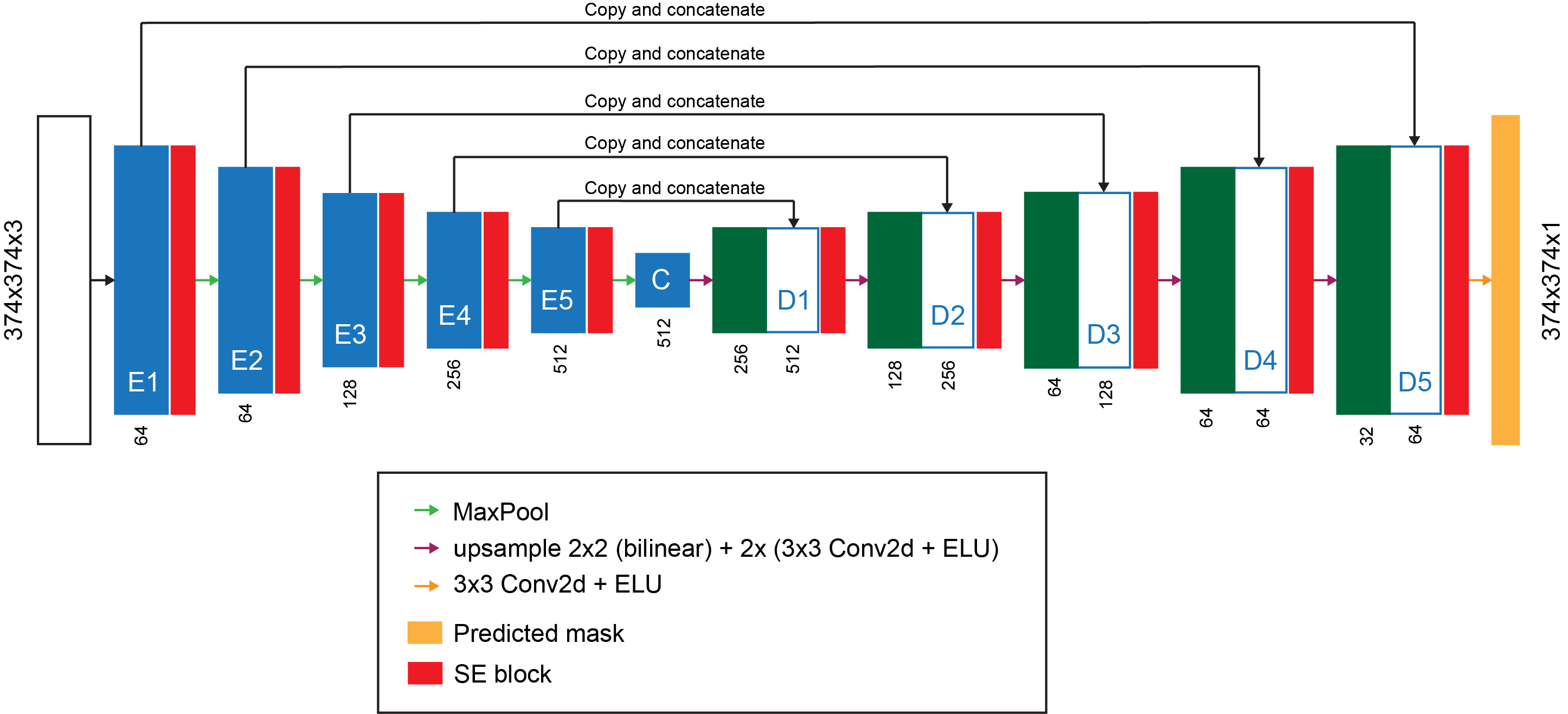}
\caption{Network Architecture with Resnet-34 as the encoder. The blocks E1-E5 are loaded with pretrained weights from Imagenet. The input is passed through the encoder layers followed by squeeze and excitation (SE) blocks before the pooling operation. SE blocks are added to this architecture as suggested in the original paper. This allows for re-calibration of the learned features adaptively. The input image is up-sampled by using bilinear interpolation at each step of the decoder and concatenated with the skip-connection input. The final output is of the same size as the input to the network.}
\label{Network}
\end{figure*}

\section{Methods}

\subsection{Model architecture and loss function}

We chose U-Net architecture for our task which is lightweight compared to the more recent state of the art models like Deeplab v3+ \cite{b17} and PSPNet \cite{b18} which exploit global context at different scales. U-Net works well where the data is limited. A typical U-Net consists of a contracting path and an expanding path. The contracting path is known as the encoder and is used to capture the contextual information from the input. The contracting branch consists of convolution operations followed by pooling operations. This series of convolution operations followed by pooling downsamples the image. In the decoder, the image is upsampled progressively followed by a convolution layer. This helps in gradually regaining the size of input to the original size. In order to be able to localize the objects detected in the image, the network needs spatial information about the image. The U-Net architecture uses skip-connections to combine the high resolution from the encoder. The output of the model also known as logits is passed through a softmax layer to generate the predictions.  
To compare the performance of our model we used a baseline model we use the vanilla U-Net architecture as proposed in the original paper. We then train our model with  Resnet-34 as the encoder of the U-Net. Slight modifications are made to the Resnet-34 architecture to help it perform better at this task. Most notably we set the $stride = 1$ for the first convolutional layer from 7x7 to 3x3. This enabled us to get a feature map of comparatively larger size, enabling the model to perform better at segmentation. We also make use of spatial and channel squeeze and excitation blocks which has shown to increase the performance of the model by quite a margin with negligible increase in the computation cost. The blocks were added in the architecture as suggested in in the original paper. We also make a few minor but vital changes. Mainly we replace  ReLU activation function with with ELU. The network architecture is as shown in Fig. \ref{Network} .Also we use Synchronized Batch Normalization \cite{b8} which allows batch norm layers to communicate with each other in multi-GPU training. We initialize the weights of the model with pretrained weights from imagenet for available layers and the rest of the layers with He Normal \cite{b9} weight initialization.

The cross-entropy loss has a simple gradient with respect to logits which make is easier to update during back-propagation. But for tasks like segmentation where the objective is to directly optimize the IoU, it's a common practice to use cross-entropy loss in combination with loss which helps in optimizing IoU \cite{b1}.

Hence we decided to use both Jaccard loss and Lovasz-Softmax \cite{b10} loss in combination with cross entropy loss.

\begin{equation}
    J(A,B) = \frac{|A \cap B|}{|A \cup B|}
\end{equation}

This equation is for discrete objects and can be extended for continuous objects as follows:

\begin{equation}
    J = \frac{1}{n}\sum_{c=1}^2 \sum_{i=1}^n (\frac{y_i^c \hat{y}_i^c}{y_i^c + \hat{y}_i^c - y_i^c \hat{y}_i^c})
\end{equation}
where $\hat{y}_i^c$ and $y_i^c$ are the corresponding predicted and ground-truth labels. 

The final loss function for this task if given as follows:
\begin{equation}\label{finalloss}
    L = \alpha J - (1 - \alpha)CE
\end{equation}
Where $J$ is Jaccard loss and $CE$ is cross-entropy. For using Lovasz-Softmax, Jaccard loss is replaced with corresponding term of Lovasz loss as given in \eqref{finalloss}. The weight of alpha is found via grid search and is set to 0.7.

\subsection{Preprocessing, training and mask generation}
We preprocessed the input by scaling the 8-bit data [0-255] to floating point values [0-1], subtracted the mean [0.485, 0.456, 0.406] from the inputs and divided by the standard deviation [0.229, 0.224, 0.225]. The numbers are the same for all Resnet based architectures trained on ImageNet.

We downscaled the input image size by a factor of 2 and trained the network. We applied heavy augmentation to our training data like all the dihedral group ($D8$) transformations, colour, brightness and contrast augmentations along with gamma correction.
Another technique that we used to is smart scale crop. Since the dataset is heavily imbalanced, using random crops might cause the imbalance to increase further. We implemented the scale-crop algorithm in such a manner that if the label mask consists of ships, some part of the ship class will be included in the crops. This way we can reduce the computation with cropping without hurting the model performance.

For any deep learning system, the devil is in the detail. For the purposes of training, we used 2 GTX-1080 GPU's. We kept the batch size constant at 8 throughout the experiments. For sampling the dataset we divided the dataset into 10 classes stratified according to the number of ships in the images. To negate the effect of class imbalance, we sample the images such that each class is visited once every C iterations as proposed in \cite{b1}.
We used AdamW \cite{b11} optimizer which fixes the weight decay issue in Adam optimizer and helps the model converge quicker. We let the model train for 60 epochs using step learning rate scheduler which decreased the learning rate from 1e-3 to 1e-5 by 0.1 every 20 epochs. Then we switched to stochastic weight averaging \cite{b13}. We train the model for 36 epochs with SGDR \cite{b14} learning rate scheduler which cycles the learning rate between 1e-5 to 1e-7. We set the cycle length as 6 and average the weights every 6 epochs. This helps in increasing the validation mIoU of our model. We also tried training the model with a 768x768 input image size. This doesn't have any significant change on the end results. Finetuning the model with for a few epochs with 768x768 image size which improved our final score by 0.4.

For validation, we use test-time augmentation like rotations in the multiples of 90$^\circ$ or horizontal and vertical flips. We average the prediction by taking their arithmetic mean. This helps in reducing the variance of our predictions. We also tried taking the geometric mean but without much difference in the results.

\section{Results}

\begin{figure*}
\centering
\captionsetup{justification=centering}
\includegraphics[width=\textwidth]{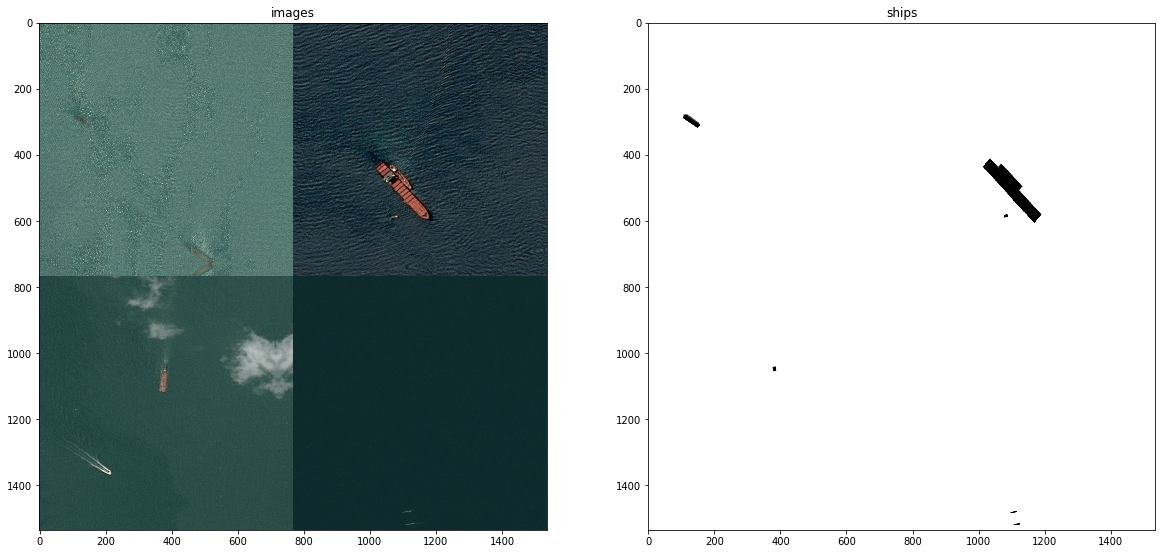}
\caption{Sample segmentation results, left: tile of 4 images, right: predicted masks for those images}
\label{results}
\end{figure*}

We present out results in table [1]. We compare our results with different loss functions as well. Some examples of our results are shown in Fig. \ref{results}.
\begin{table}[h!]
\begin{center}
    \begin{tabular}{|c|c|c|c|}
    \hline
    Encoder Network & Params Count (millions) & Jaccard & Lovasz \\
    \hline \hline
    VGG-19 & 143M & 0.796 & 0.772 \\
    Resnet-34 & 23.2M & 0.832 & \textbf{0.845} \\
    Resnet-50 & 25.6M & \textbf{0.834} & 0.826 \\
    \hline
    \end{tabular}
    \end{center}
    \caption{Mean F2-score on local validation set for different encoders and loss functions}
    \label{table :1}
\end{table}

\begin{table}[h!]
\begin{center}
    \begin{tabular}{|c|c|c|}
    \hline
    Network Architecture & SE-Block & F2-score  \\
    \hline \hline
    U-Net w/ Resnet-34 & No & 0.827\\
    SE U-Net w/ Resnet-34 & Yes & \textbf{0.845} \\
    \hline
    \end{tabular}
    \end{center}
    \caption{Mean F2-score on local validation set for U-Net with and without SE Block}
    \label{table :2}
\end{table}

We compare results of Resnet-34 with other encoders like Resnet-50 and VGG-19. The validation score of Resnet based models ate higher as compared to VGG-19. This is because of a better gradient flow due to skip-connections. Our heaviest model Resnet-50 gives an inferior performance as compared to Resnet-34. This can be attributed to two reasons. i). The presence of spatial and channel squeeze and excitation blocks in Resnet-34 which improves the ability of the model to learn channel and spatial inter-dependencies. ii) Lovasz-Softmax requires good gradient flow and this gets tougher as the size of the architecture keeps on increases. Resnet-34 outperforms all other models when trained using Lovasz-Softmax loss.

On performing ablation study on the model by training with and without squeeze and excitation block, it's clear that it give a significant boost in performance for segmentation models.

For comparing the speed of model, we compare the results against popular segmentation PSPNet with a Resnet-50 backbone and DeepLab v3+ with aligned modified Xception as backbone.

\begin{table}[h!]
\begin{center}
    \begin{tabular}{|c|c|c|}
    \hline
    Network Architecture & Inference Speed (Frames/second) & F2-score  \\
    \hline \hline
    PSPNet-50 & 0.19 & \textbf{0.862}\\
    DeepLab v3+ & 0.27  &  0.855\\
    SE U-Net & \textbf{1.21} & 0.845 \\
    \hline
    \end{tabular}
    \end{center}
    \caption{Inference speed and mean F2-score performance on validation set.}
    \label{table :3}
\end{table}

PSPNet and DeepLabv3+ give higher performance as compared to our method, this can be attributed to the use of multi-scale features by dedicating a separate module to it. Our method gives comparable performance and a 6x speed upgrade as compared to the top performing method, i.e. PSPNet.

\section{Conclusion}
In this paper, we presented an approach to ship segmentation in satellite imagery. We used U-Net architecture with Resnet-34 as the backbone. The quality of segmentation is improved because of three factors. i). Using SE blocks in the architecture. ii). Using Lovasz-Softmax to optimize the mIoU score. iii) Using smart-cropping augmentation to tackle data imbalance. Some areas of improvement for future work would be producing fine-grained masks. Our model produces a single mask for ships which are very close to each other. This could be improved by applying post-processing like watershed transformation. We chose not to apply additional-postprocessing to keep the model computationally efficient.

\section*{Acknowledgment}

The authors would like to thank the department of computer science of SRM Institute of Science and Technology for providing us with useful suggestions for conducting these experiments.

\end{document}